\RequirePackage{fix-cm}
\RequirePackage{ifluatex}

\documentclass[smallcondensed]{svjour3}     
\smartqed  
\usepackage{geometry}
\usepackage{blindtext}
\usepackage[numbers,sort&compress]{natbib}

\usepackage[driverfallback=dvips]{hyperref}

\usepackage{indentfirst}

\usepackage{graphicx}

\usepackage{multirow}

\usepackage{amsmath,bm}

\usepackage{amsfonts,amssymb}
\usepackage{color}

\usepackage{algorithm}
\usepackage{algpseudocode}
\usepackage{subfig}
\usepackage{makecell}

\begin{document}

\begin{sloppypar}

\title{Reentry Risk and Safety Assessment of Spacecraft Debris Based on Machine Learning}

\author{Hu Gao$^{a}$\and
	Zhihui Li$^{b,c}$\and
	Depeng Dang$^{a,*}$\and
	Jingfan Yang$^{a}$\and Ning Wang$^{a}$
}

\institute{
	$^a$ School of Artificial Intelligence, Beijing Normal University, Beijing 100000, China\\
	$^b$ China Aerodynamics Research and Development Centre, Mianyang 621000, China\\
	$^c$ National Laboratory for Computational Fluid Dynamics, BUAA, Beijing 100191, China \\
	*Corresponding Author\\
	 \at
	Hu Gao\at
	\email{gaoh@mail.bnu.edu.cn}
	\and
	Zhihui Li \at
	\email{zhli0097@x263.net}
	\and
	Depeng Dang\at
	\email{ddepeng@bnu.edu.cn}
	\and
	Jingfan Yang\at
	\email{yangjingfan@mail.bnu.edu.cn}
	\and
	Ning Wang\at
	\email{wangningbnu@mail.bnu.edu.cn}
}
\date{Received: date / Accepted: date}
\maketitle
\begin{abstract}
Uncontrolled spacecraft will disintegrate and generate a large amount of debris in the reentry process, and ablative debris may cause potential risks to the safety of human life and property on the ground. Therefore, predicting the landing points of spacecraft debris and forecasting the degree of risk of debris to human life and property is very important. 
In view that it is difficult to predict the process of reentry process and the reentry point in advance, and the debris generated from reentry disintegration may cause ground damage for the uncontrolled space vehicle on expiration of service. In this paper, we adopt the object-oriented approach to  consider the spacecraft and its disintegrated components as consisting of simple basic geometric models, and introduce three machine learning models: the support vector regression (SVR), decision tree regression (DTR) and multilayer perceptron (MLP) to predict the velocity, longitude and latitude of spacecraft debris landing points for the first time. Then, we compare the prediction accuracy of the three models. Furthermore, we define the reentry risk and the degree of danger, and we calculate the risk level for each spacecraft debris and make warnings accordingly.  The experimental results show that the proposed method can obtain high accuracy prediction results in at least 15 seconds and make safety level warning more real-time.
\end{abstract}

\section{Introduction}
\label{intro}
In the process of uncontrolled atmospheric reentry, spacecraft disintegrate under a variety of forces, such as gravity and aerodynamics; and produce a large amount of debris~\cite{txh, shp2}.  Most debris is destroyed by thermal ablation with the atmosphere, but some debris may survive and impact the ground, posing a threat to the safety of people. If debris crashes into buildings or in densely populated areas, unimaginable grave consequences will occur~\cite{shp2, tl}. Therefore, improving the precision and efficiency of spacecraft debris risk assessment and accurately predicting the landing time, location and risk level of debris is very important~\cite{hcj}. 

The traditional spacecraft debris reentry risk assessment approach is based on physical modelling and involves different modules, such as the trajectory, aerodynamic/aerothermal characteristics, ablation, disintegration, ground risk assessment, and their coupling. In order to predict the reentry trajectory of dismantled spacecraft fragments, aerospace professionals need to use orbital dynamics and hydrodynamics to model and analyse multiscale complex flow problems and then obtain landing points by solving the reentry trajectory~\cite{lj}. The above approach requires the Monte Carlo method, Gas-Kinetic Unified Algorithm (GKUA) for solving the Boltzmann model equations and other probabilistic methods to simulate molecular collisions in the atmosphere. Due to the high complexity and limitation of computing power and energy consumption, the traditional approach takes too much time, and the accuracy is insufficient.

In the past 20 years, advances in machine learning have driven the development of many disciplines. Machine learning is used in many fields, such as medicine and finance, to predict the probability of events or future trends. Researchers in some safety-critical fields, such as aerospace, have also started to use machine learning to solve problems. For example, ~\cite{okj} use an recurrent neural network (RNN) to predict re-entry trajectories of uncontrolled space objects. ~\cite{ss} use a machine learning approach for spacecraft break-up predictions. However, due to the rapid development of machine learning, many new machine learning approaches are not used in aerospace. Therefore, the prediction of spacecraft reentry debris landing points based on machine learning has important scientific and engineering value.

In this paper, we propose a risk safety assessment approach for spacecraft debris based on machine learning.  We adopt the object-oriented approach to  consider the spacecraft and its disintegrated components as consisting of simple basic geometric models, and introduce three machine learning algorithms to predict the velocity, longitude and latitude of spacecraft debris landing points for the first time. Then, we define the reentry risk and the degree of danger. Furthermore, we calculate the risk level for each spacecraft debris and make warnings accordingly.  Compared with those existing works, our main contributions are as follows:

\begin{enumerate}
	\item  By using machine learning, there is no need to consider the physical process of spacecraft debris reentry. From the input of raw data to the output of task results, the whole training and prediction process is completed in the model. Our approach is also the only approach based on machine learning for spacecraft debris risk safety assessment.
	
	\item We compares the accuracy of three machine learning regression algorithms, support vector machine, decision tree and multi-layer perceptron, to predict the landing location of spacecraft debris, and analyzes the reasons for such results. 
	
	\item We redifine the reentry risk and the economy, population and kinetic energy of the debris landing points are taken as risk factors. Then, according to the law of Bradford, we divided the debris risk into 5 equal parts with a step size of 0.2, and formulated the degree of debris risk.
	
	\item The experimental results show that the proposed method in our paper can obtain high acccuracy prediction results in at least 10 seconds. Among them, by using decision trees, the average longitude prediction error is $0.96^{\circ}$ , the average latitude prediction error is $0.53^{\circ}$, and the average velocity prediction error is $0.008m/s$. 
\end{enumerate}

\section{Related Work}
\label{sec:1}

\subsection{Space debris reentry risk assessment}The research on space debris reentry prediction and ground risk assessment started ten years ago, and mature software systems have been developed. The common approaches for spacecraft debris reentry prediction, which can be divided into object oriented and spacecraft oriented, require a strong simulated physical model to determine the trajectory. The object-oriented method first reduces the shape of the spacecraft and its components to basic geometries and then simulates the reentry process by modelling aerodynamic, aerothermal and ablation processes. The spacecraft-oriented method directly models the real shape of spacecraft and calculates the aerodynamic and aerothermal properties. The main space reentry prediction software includes DAS, ORSAT, SCARAB, ORSAT, DRAMA/SESAM and DEBRISK~\cite{das,bsa,fb,shp2,shp,tl,tl2,bf,ja,ipf}. DAS and ORSAT are based on object-oriented methods, and SCARAB is based on a spacecraft-oriented method. ~\cite{lj} used the Monte Carlo method to simulate aerodynamic and aerothermal characteristics and then analysed the spacecraft disintegration process. On the basis of the dynamics model, ~\cite{txh} propose a spacecraft landing point prediction approach based on upper wind real-time correction. ~\cite{shp} use space debris considering byproduct generation, analyse the surviving space fragments and then assess ground risk. However, the above approach need to simulate molecular collisions in the atmosphere. Due to the high complexity and limitation of computing power and energy consumption, this methods take too much time, and the accuracy is insufficient.

\subsection{Machine learning:} Machine learning is a type of approach that enables computer to "learn" by simulating the human brain. Machine learning can acquire the potential laws from data and use the laws to predict unknown data. The application of machine learning in aerospace can be traced back to the 1970s. Voyager I, which was launched by NASA in 1977, applies expert systems containing command decoding, fault detection and correction. In the 21st century, deep learning, a branch of machine learning, has made breakthroughs; therefore, in many fields, work based on machine learning has attracted researchers again. ~\cite{gkb} use machine learning approaches such as Tar3 and the naive Bayes classifier to predict possible influencing factors in the spacecraft reentry process. ~\cite{lyt} use deep learning to establish an adaptive spacecraft situation analysis system. ~\cite{ss} apply a machine learning method to spacecraft fracture prediction and use a random forest regression algorithm and the data simulated by SCARAB as the dataset.
 However, for the data with different value attributes, the attributes with more value have a greater impact on the random forest, and the generated attribute weights are not credible. Furthermore, the random forest has been proven to be overfitted in some high noise level regression problems. In this paper, We compares the accuracy of three machine learning regression algorithms, support vector machine, decision tree and multi-layer perceptron, to predict the landing location of spacecraft debris, and analyzes the reasons for such results. And then we calculate the risk level for each spacecraft debris and make warnings accordingly.
\subsection{Definition of the degree of reentry risk:}
~\cite{hrf} define the degree of reentry risk as the number of people on the ground who may be affected by spacecraft debris, and the formula is the following: ${E=\sum_{i=1}^{n} \rho_{i}A_{i}}$, where ${\rho_{i}}$ is the population density of the landing point, ${A_{i}}$ is the cross-sectional area of debris, $i$ is the $i_{th}$ piece of debris, and $n$ is the number of pieces of spacecraft debris. On this basis, our paper considers the economy where debris falls and the kinetic energy when debris falls.

\section{Methodology}
\label{sec:model2}

In this section, we introduce the following contents: (1) The velocity, longitude and latitude prediction of spacecraft debris landing points based on a machine learning regression; and (2) Risk assessment and the degree of dangerous based on the economy, population and debris kinetic energy of landing points.

\subsection{Landing point velocity, longitude and latitude prediction}
\begin{figure}[!htb] 
	\centering
	\includegraphics[width=0.75\textwidth]{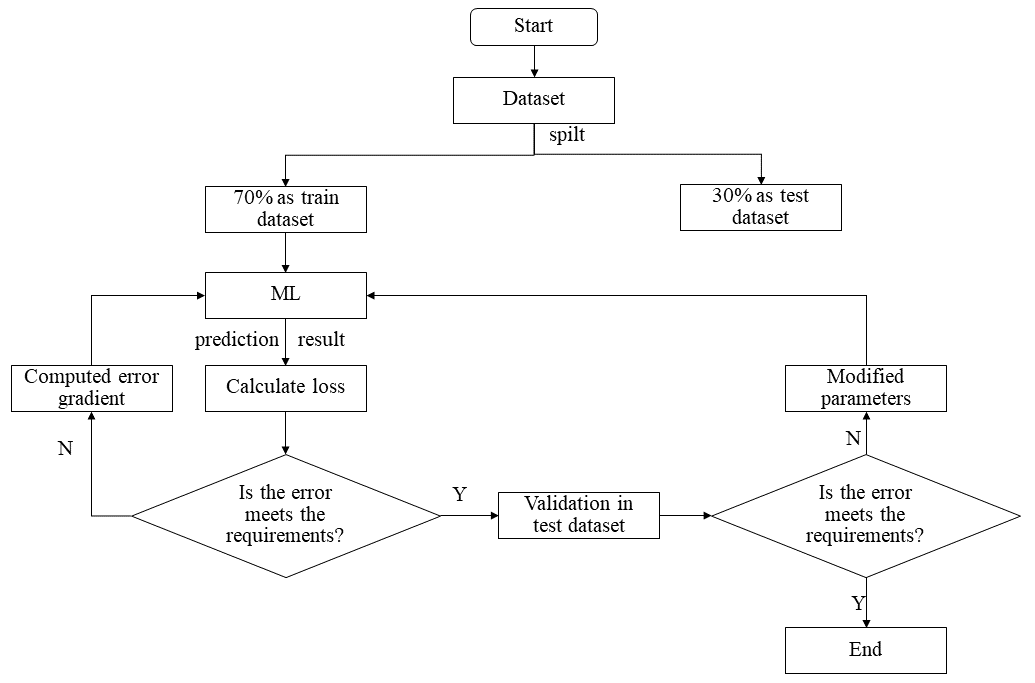}
	\caption{The machine learning training process}
	\label{fig:001}
\end{figure}

The velocity, longitude and latitude prediction of spacecraft debris landing points can be considered a regression problem, and we use three machine learning algorithms to solve the problem: the support-vector regression, decision tree and multilayer perceptron. The machine learning training process is shown in Fig.~\ref{fig:001}. First, we preprocessed the data to make a dataset, the dataset is $D = \{(X_1, Y_1), (X_2, Y_2),..., (X_n, Y_n)\}$, where $X$ is the input variable, $X_i = (x^{(1)}_{i}, x^{(2)}_{i},..., x^{(n)}_{i})$ is the feature vector, $Y$ is the output variable; $n$ is the number of features, $i=1,2,...,N$, and $N$ is the sample size. And next divided it into a training set and a test set. Then, we put the training dataset  into the machine learning model for training, and calculate the loss of the predicted value and the target value of the model according to the loss function. Next, we judge whether the error meets the requirements, if "N", we continue training the model. Otherwise, we test the predictive power of the model on the test dataset. If the requirements are met,  the model training completed. If not, we modify the model parameters and retrain it.

\subsubsection{Support Vector Regression}
A support vector machine (SVM) is a method originally used for classification problems and can be used for regression problems after improvement, namely, the support vector regression (SVR)~\cite{zx}. The goal of an SVR is to find a regression plane so that the predicted value is as close to the regression plane as possible. The approximate continuous valued function of an SVR from a geometric view is as follows:
\begin{equation}
	\label{equ:1}
	\hat{y}=f(x)=<w,x>+b=\sum_{j=1}^{M}w_{j}x_{j}+b,~~~~~~y,b \in R,~~x,w\in R^M
\end{equation}
\begin{equation}
	\label{equ:2}
	f(x)=\begin{bmatrix}
	w\\b
	\end{bmatrix}^T 
	\begin{bmatrix}
	x\\1
	\end{bmatrix}
	=w^Tx+b,~~~~~~x,w \in R^{m+1}
\end{equation}

An SVR reduces the above function approximation problem to an optimisation problem, then finds a hyperplane-centric interval zone, and finally minimises the prediction error (the distance between the predicted value and the target value). The interval zone is defined as:
\begin{equation}
	\label{equ:3}
	\min w = \frac{1}{2}||w||^2
\end{equation}

where $||w||$ is the size of the approximation plane normal vector.

\begin{figure}[!htb] 
	\centering
	\includegraphics[width=0.5\textwidth]{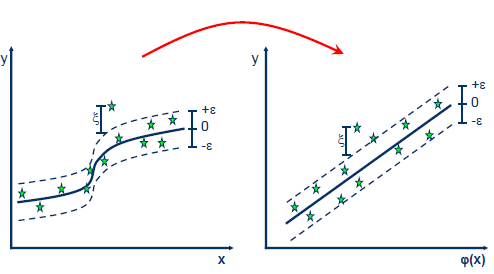}
	\caption{The Nonlinear SVR. The original space of the input data is mapped to a higher dimensional feature space}
	\label{fig:xfig1}
\end{figure}

An SVR has two main advantages: one advantage is that the computational complexity no longer depends on the input space dimension, and the other is that the model quality depends on the correct setting of parameters such as the kernel function. In this paper, we select the nonlinear SVR and use a  radial basis kernel function as the Eq.~\ref{equ:4} to put the data into a high-dimensional feature space, then fit the optimal linear plane. As shown in Fig.~\ref{fig:xfig1}.  
\begin{equation}
	\label{equ:4}
	 K(x,x^{\prime})=exp(-\frac{||x-x^{\prime}||^{2}}{2 \sigma ^ 2})
\end{equation}

Finally, the objective function of our model is defined as Eq.~\ref{equ:5}. The smoothest plane is sought by minimizing the square of the vector norm $w$, and the error of the predicted value for each training data is at most equal to $\epsilon$. To allow for outliers, data with prediction errors greater than $\epsilon$ are penalized using an $\epsilon-$ sensitive loss function. The constrained optimization problem is then reformulated into a dual problem form using lagrange multipliers, and for each constraint, quadratic programming is used to determine, after which the deviation of the best weight is calculated, and then the predicted value $\hat{y}$ is given.
\begin{equation}
	\label{equ:5}
	\hat{y}=\sum_{i=1}^{N} w_{i} K(x,x^{\prime})+b
\end{equation}

\subsubsection{Decision Tree Regression}
The decision tree regression (DTR) is a prediction model based on a set of binary rules, it divides the regions, and each region corresponds to a uniform predicted value.  Each individual decision tree is a simple model with branches, nodes and leaves.

A decision tree uses a heuristic method to divide the feature space. In each division, the possible values of all features in the current dataset are examined, and 
the best feature according to the lowest mean square error is selected as the splitting criterion~\cite{mc,mx,ssr,gkf}. The decision tree algorithm ~\ref{alg:decision_tree} is described in detail below. As algorithm ~\ref{alg:decision_tree} shows, the $j_{th}$ feature variable $x^{(j)}$ and its value $s$ in the training set are taken as the splitting variable and splitting point, respectively. Then, two regions $R_{1}(j, s) = \{x|x^{(j)} \leq s\}$ and $R_{2}(j, s) = \{x|x^{(j)} > s\}$ are defined. Finally, j and s are searched, which can minimize the square sum of errors of two regions according to the Eq~\ref{equ:6}, and each region corresponds to a uniform predicted value.
\begin{equation}
	\label{equ:6}
	\min[^{min}_{c1}\sum_{x_{i} \in R_{1}(j,s)}(y_{i}-c_{1})^{2}) + ^{min}_{c2}\sum_{x_{i} \in R_{2}(j,s)}(y_{i}-c_{2})^{2})]
\end{equation}

After splitting, $c_1$ and $c_2$ are fixed output values, which are the averages of $Y$ in their regions. Therefore, Eq.~\ref{equ:6} can be rewritten as:
\begin{equation}
	\label{equ:7}
	\min[\sum_{x_{i} \in R_{1}(j,s)}(y_{i}- \hat c_{1})^{2}) + \sum_{x_{i} \in R_{2}(j,s)}(y_{i}- \hat c_{2})^{2})]
\end{equation}

where\begin{equation}
	\begin{aligned}
		\label{equ:8}
		\hat{c_{1}}=\frac{1}{N_{1}} \sum_{x_{i} \in R_{1}(j,s)} y_{i}
		\\
		\hat{c_{2}}=\frac{1}{N_{2}} \sum_{x_{i} \in R_{2}(j,s)} y_{i}
	\end{aligned}
\end{equation}

\begin{algorithm}[!htb]
	\caption{Decision Tree Regression}
	\label{alg:decision_tree}
	\hspace*{0.02in}{\bf Input:}
	Training dataset D\\
	\hspace*{0.02in}{\bf Output:}
	Decision tree f(x)
	\begin{algorithmic}[1]
		\State $f(x) = null$
		\While{$j \neq null$}
		\While{$s \neq null$}
		\If{Eq.~\ref{equ:6}}
		
		\State $ \hat{c_{m}}=\frac{1}{N_{m}} \sum_{x_{i} \in R_{m}(j,s)} y_{i} $
		
		\EndIf
		\EndWhile
		\EndWhile
		\State Pruning
		\State Divide dataset D into M regions $R_1 , R_2 , ... , R_M$
		\For{m=1;m<M;m++}
		\If{$x \in R_{m}$}
		\State $f(x)=f(x)+\hat{c_{m}} $
		\EndIf 
		\EndFor
		\State return f(x)
	\end{algorithmic}
\end{algorithm}

The generation of the decision tree completely depends on the training sample, and the decision tree can fit the training samples perfectly. However, such a decision tree is too large and complex for the test sample and may produce a high classification error rate, namely, overfitting.
Therefore, it is necessary to simplify a complex decision tree and remove some nodes to solve overfitting, which is called pruning.

Pruning has two components: prepruning and postpruning. Prepruning terminates the growth of decision trees early in the decision tree generation process to avoid the presence of too many nodes. Prepruning is simple but not practical because it is difficult to determine when an algorithm should stop. Postpruning replaces a low-confidence node subtree with a leaf node after decision tree construction, and the leaf node is labelled with the highest frequency class in the subtree. There are two postpruning methods: one method divides the training dataset into a growth set and a pruning set, and the other method uses the same dataset for growth and pruning. Common postpruning methods are cost complexity pruning (CCP), reduced error pruning (REP), pessimistic error pruning (PEP) and minimum error pruning (MEP)~\cite{whn}. 

\begin{figure}[!htb]
	\centering
	\subfloat[]{\label{fig:xfig0_1}
		\includegraphics[width=0.95\textwidth]{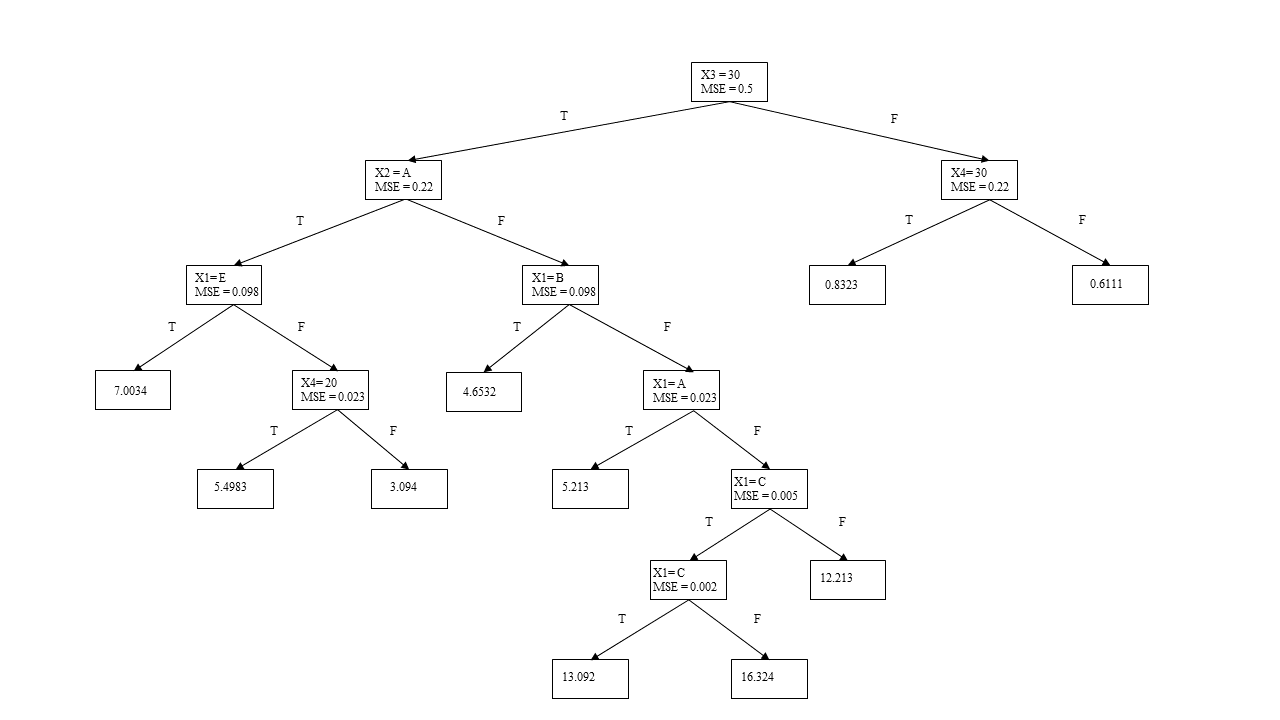}}
	\quad
	\subfloat[]{\label{fig:xfig0_2}
		\includegraphics[width=0.95\textwidth]{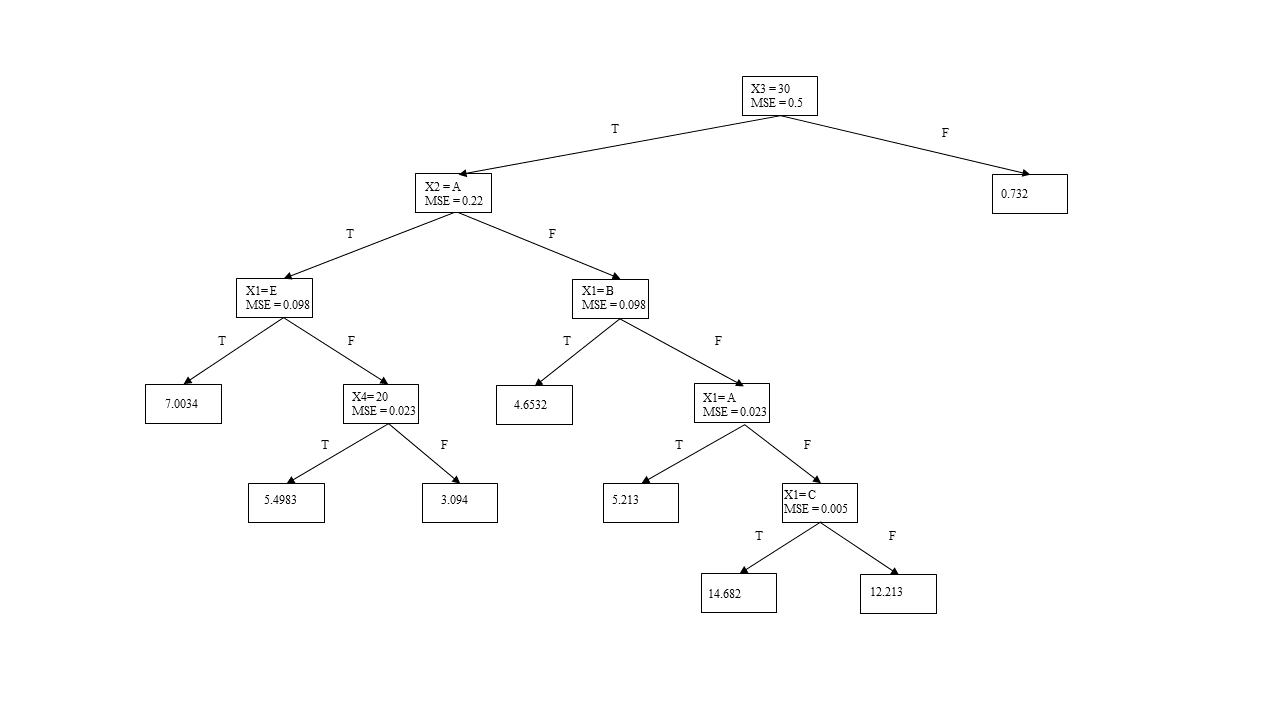}}
	\caption{The decision regression tree. Figure (b) shows the figure (a) after pruning.}
	\label{fig:xfig0_0}
\end{figure}

The pruning process is shown in the Fig.~\ref{fig:xfig0_0}. All the subtrees are traversed from bottom to top.For a leaf node that covers $n$ samples with $e$ errors, the error rate is $(e + 0.5) / n$, where $0.5$ is the penalty factor. For a subtree with $L$ leaf nodes, the misjudgement rate is defined as:
\begin{equation}
	\label{equ:9}
	ErrorRatio = \frac{\sum_{i=1}^{L} e_{i}+0.5L}{\sum_{i=1}^{L}n_{i}}
\end{equation}

where $e_i$ is the number of samples of the $i_{th}$ leaf node misclassification, and $n_i$ is the total number of samples of the $i_{th}$ leaf node. 

If a subtree is misclassified with a sample value of 1 and correctly classified with a sample value of 0, the number of subtree misjudgements obeys a Bernoulli distribution. Therefore, the mean value and standard deviation of the number of subtree misjudgements can be obtained as follows:
\begin{equation}
	\label{equ:10}
	ErrorMean = ErrorRatio * \sum_{i=1}^{L}n_i
\end{equation}
\begin{equation}
	\label{equ:11}
	ErrorSTD = \sqrt{ErrorRatio * \sum_{i=1}^{L}n_{i}*(1-ErrorRatio)}
\end{equation}

After replacing the subtree with a leaf node, the misjudgement rate of the leaf node is:
\begin{equation}
	\label{equ:12}
	ErrorRatio^{\prime} = \frac{e^{\prime}+0.5}{n^{\prime}}
\end{equation}

where $e^{\prime}=\sum_{i=1}^{L}e_{i}$ , and $n^{\prime}=\sum_{i=1}^{L}n_{i}$ .
Furthermore, the number of misjudgements of this leaf node also obeys a Bernoulli distribution, so the mean number of misjudgements of this leaf node is:
\begin{equation}
	\label{equ:13}
	ErrorMean^{\prime} = ErrorRatio^{\prime} * n^{\prime}
\end{equation}

The pruning condition is given as Eq.~\ref{equ:14}. When the pruning condition is satisfied, the subtree will be replaced by a leaf node.
\begin{equation}
	\label{equ:14}
	ErrorMean + ErrorSTD \ge ErrorMean^{\prime}
\end{equation}

Finally, the data is input into the constructed decision regression, and the optimal region is selected according to the rules of the decision regression tree, and the final predicted value is obtained.

\subsubsection{Multilayer Perceptron}
\begin{figure}[!htb] 
	\centering
	\includegraphics[width=0.6\textwidth]{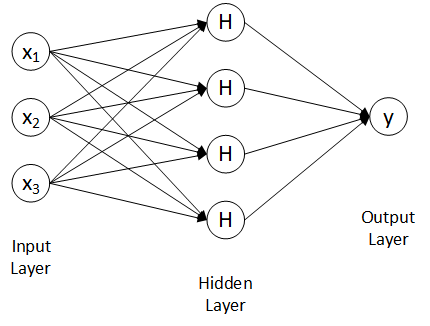}
	\caption{The network structure of MLP. On this figure, we take three input variables as examples and output the final predicted value $\hat{y}$ through a layer of perceptron}
	\label{fig:xfig2}
\end{figure}

Neural networks were originally developed by researchers to imitate the neurophysiology of the human brain. As shown in Fig.~\ref{fig:xfig2}, a multilayer perceptron (MLP) is an artificial neural network model that is composed of at least three layers: an input layer, a hidden layer and an output layer. 
In addition to the input layer nodes, each node is a neuron activated by a nonlinear activation function~\cite{ly, jgp}.

In the MLP learning process, the feature vector $X$ is input into the model for forward propagation to obtain the predicted value $\hat{y}$. This process can be defined as Eq.~\ref{equ:15}, where $W_1$ and $W_2$ are the weight matrices of the first and second layers, respectively; $f( \bullet )$ is the ReLU activation function and $f(H) = max(0,H)$~\cite{bk}. Then, the difference between the prediction value and target value is calculated, as shown in Eq.~\ref{equ:16}.
\begin{equation}
	\label{equ:15}
	\begin{aligned}
		H = XW_{1} + B_{1}
		\\
		\hat{y} = f(H)W_{2} + B_{2}
	\end{aligned}
\end{equation}
\begin{equation}
	\label{equ:16}
	e_{j} =   y_{j} - \hat{y_{j}}
\end{equation}

where $y_{j}$ and $\hat{y_{j}}$ are the target value and prediction value of node $j$, respectively.

\begin{figure}[!htb] 
	\centering
	\includegraphics[width=0.5\textwidth]{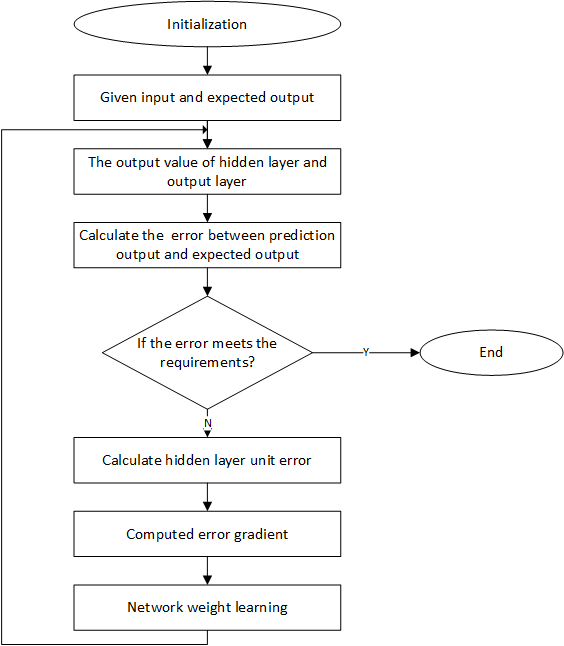}
	\caption{The flowchart of an MLP}
	\label{fig:xfig3}
\end{figure}

In backpropagation, the weights matrices  $W_1$ and $W_2$ are updated to minimize the overall output error, and the partial derivative of the objective function with respect to each neuron weight is calculated layer by layer. This results in the gradient of the objective function with respect to the weight vector, which is used as the basis for updating the weights, as shown in Eq.~\ref{equ:17}, and the $\alpha$ is the momentum factor. The learning is completed in the process of updating the weights to make the error reach the expected value. The MLP flowchart is shown in Fig.~\ref{fig:xfig3}.
\begin{equation}
	\label{equ:17}
	\begin{aligned}
		&\xi = \frac{1}{2} \sum_{j} e_{j}^{2}
		\\
		&\frac{\partial \xi}{\partial W_{1}} = \frac{\partial \xi}{\partial \hat{y}}
		\frac{\partial \hat{y}}{\partial f(H)}
		\frac{\partial f(H)}{\partial H}
		\frac{\partial H}{\partial W_{1}}
		\\
		&W_{i} = W_{i} - \alpha \frac{\partial \xi}{\partial W_{i}}
	\end{aligned}
\end{equation}

\subsubsection{Loss function}
In order to train the model proposed in this paper, we use the mean square error (MSE) as the loss function. MSE  is the most commonly used regression loss function, which refers to the sum of the squared distance between the target value and the predicted value, as shown in Eq.~\ref{equ:08}, where $y_i$ is the target value, $\hat{y_i}$ is the predicted value by the machine learning algorithm, and $n$ is the sample size.

\begin{equation}
	\label{equ:08}
	MSE = \frac{\sum_{i=1}^{n} (y_i - \hat{y_i})^2}{n}
\end{equation}

\subsection{Risk assessment}
\subsubsection{Calculation of the Economy, Population and Debris Kinetic Energy of Landing Points}
The object oriented method can be used to abstract the fragments into finite groups with simple geometric shapes, and ballistic simulation can be carried out for each group of fragments. Groups with similar landing points can be integrated. Meanwhile, it is noted that the ratio of surface to mass of fragments has a great influence on the ballistic characteristics, so the influence of fragment size should be considered after the geometric grouping scheme is determined.  The final fragment grouping scheme is shown in Table~\ref{tab:table001}.
\begin{table}[h]
	\centering 
	\caption{Fragment quality scheme.}
	\label{tab:table001}
	\renewcommand\arraystretch{1.5}
	\begin{tabular}{|c|c|c|c|}
		\hline
		
		\makecell[c]{Pieces of type} 
		&\makecell[c]{Scale of 1}
		&\makecell[c]{Scale of 2}
		&\makecell[c]{Scale of 3}
		\\ \hline
		\makecell[c]{A (Block: Round ball)} 
		&\makecell[c]{$W_{A1}$}
		&\makecell[c]{$W_{A2}$}
		&\makecell[c]{$W_{A3}$}
		\\ \hline
		\makecell[c]{B (Block: cuboid)} 
		&\makecell[c]{$W_{B1}$}
		&\makecell[c]{$W_{B2}$}
		&\makecell[c]{$W_{B3}$}
		\\ \hline
		\makecell[c]{C (Flake)} 
		&\makecell[c]{$W_{C1}$}
		&\makecell[c]{$W_{C2}$}
		&\makecell[c]{$W_{C3}$}
		\\ \hline
		\makecell[c]{D (Rhabditiform: 
			$l=5a$)} 
		&\makecell[c]{$W_{D1}$}
		&\makecell[c]{$W_{D2}$}
		&\makecell[c]{$W_{D3}$}
		\\ \hline
		\makecell[c]{E (Rhabditiform: 
			$l=10a$)} 
		&\makecell[c]{$W_{E1}$}
		&\makecell[c]{$W_{E2}$}
		&\makecell[c]{$W_{E3}$}
		\\ \hline
	\end{tabular}
\end{table}

The smaller the single mass of a certain kind of fragment is, the larger the quantity of this kind of fragment is, and the relationship between the mass and the quantity is  linear after taking logarithms~\cite{lsw}. Therefore, we set up:

\begin{equation}
	\label{equ:001}
	n = C \cdot m^{-k}
\end{equation}

Where $C$ is determined based on quality constraints, $k$ varies according to disassembly type and is chosen as $0.553$ in this paper. 

\begin{figure}[!htb] 
	\centering
	\includegraphics[width=0.7\textwidth]{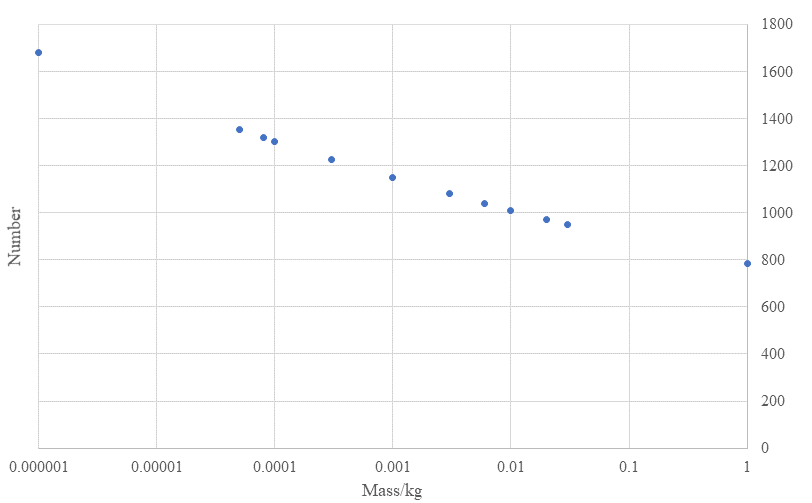}
	\caption{Relationship between quantity and mass for debris groups}
	\label{fig:xfig001}
\end{figure}

Then, combined with Table.~\ref{tab:table001} and Eq.~\ref{equ:001}, the weight of each fragment can be determined. For example: Assuming that the total mass of debris is $200kg$ and the three scales are $0.1m$, $0.01m$ and $0.001m$, the relationship between the quantity and mass of each group of debris is shown in Fig.~\ref{fig:xfig001}. Finally, the spacecraft was determined to disintegrate into seven fragments, and the mass scheme is shown in Table.~\ref{tab:table00}.

\begin{table}[h]
	\centering 
	\caption{Finite grouping scheme for debris.}
	\label{tab:table00}
	\tabcolsep=0.6cm
	\renewcommand\arraystretch{2}
	
	\begin{tabular}{|c|c|c|c|}
		\hline
		
		\makecell[c]{Debris  number} 
		&\makecell[c]{Quality(kg)}
		\\ \hline
		\makecell[c]{1} 
		&\makecell[c]{16.0}
		\\ \hline
		\makecell[c]{2} 
		&\makecell[c]{72.7}
		\\ \hline
		\makecell[c]{3} 
		&\makecell[c]{65}
		\\ \hline
		\makecell[c]{4} 
		&\makecell[c]{225.5}
		\\ \hline
		\makecell[c]{5} 
		&\makecell[c]{8.0}
		\\ \hline
		\makecell[c]{6} 
		&\makecell[c]{20.7}
		\\ \hline
		\makecell[c]{7} 
		&\makecell[c]{108.5}
		\\ \hline
	\end{tabular}
\end{table}

The NASA Socioeconomic Data and Applications Centre releases the Gridded Population of World (GPW) that provides a spatial grid population distribution layer that is compatible with social or economic datasets. The population grid layer is the administrative divisions worldwide, and the vector dataset and national digital identifier grid for each input administrative region centre point are also included in the population layer ~\cite{nasa}.

We use the velocity, longitude and latitude predicted by machine learning in the first part to calculate the economy, population and kinetic energy of spacecraft debris landing points. The calculation steps are given as follows:

\begin{enumerate}
	\item  Calculate and match the longitude and latitude of a landing point predicted in the first part with the central point position of each administrative region in the GPW, find the nearest administrative region to the landing point, and obtain the population density of the spacecraft debris landing area.

	\item 	Gather the GDP of all regions worldwide, and find the regional GDP of the administrative region obtained in step 1 from the GDP dataset to measure the economic level of the spacecraft debris landing area.
	
	\item Using the debris landing velocity and mass predicted in the first part, the kinetic energy of spacecraft debris landing is calculated according to the kinetic energy formula $KE=\frac{1}{2}mv^{2}$.
\end{enumerate}

\subsubsection{Definition of Reentry Risk}
The casualty area is used to assess the ground risk caused by spacecraft debris during reentry.
The total casualty area is calculated as the sum of the casualty areas of all $n$ pieces of debris reentering as follows:
\begin{equation}
	\label{equ:18}
	A_{c} = \sum_{i=1}^{n}(\sqrt{A_{h}}+\sqrt{A_{i}})^{2}
\end{equation}

where $A_{h}$ is the human ground projected cross-sectional area $A_{i}$ is the $i_th$ landing fragment cross-sectional area. In NASA safety standard NSS 1740.14, the human ground projected cross-sectional area is $0.36m^{2}$.
The debris landing points in our work are not the same; therefore, we discuss a single degree of debris reentry risk based on the damage of human life and property, which is defined as follows:
\begin{equation}
	\label{equ:19}
	E = A_{i} *  \rho_{i} * u_{i} *KE_{i}
\end{equation}

where $\rho_{i}$ is the population density of the landing point of the $i_{th}$ fragment, $u_i$ is the economic situation of the landing point of the $i_{th}$ fragment, $KE_i$ is the landing kinetic energy of the $i_{th}$ fragment, and $A_i$ is the cross-sectional area of the landing of the $i_th$ fragment. The degree of reentry risk is the quantitative basis for risk grading, risk grading is the theoretical basis of the degree of danger, and the degree of danger is the result of risk safety assessment.
Because there is a large magnitude difference between fragment reentry risk, it is necessary to normalise the reentry risk data to the $[0,1]$ interval. The minimum value of sample data $Y$ is $min$, the maximum value is $max$, and the calculation coefficient $k$ is:
\begin{equation}
	\label{equ:20}
	k = \frac{1}{max - min}
\end{equation}

Then, the data are normalised to the $[0,1]$ interval:
\begin{equation}
	\label{equ:21}
	norY = k * (Y-min)
\end{equation}

The normalisation operation is expressed as $nor$, and the debris risk is defined as:
\begin{equation}
	\label{equ:22}
	W = norE
\end{equation}

\begin{figure}[!htb] 
	\centering
	\includegraphics[width=0.7\textwidth]{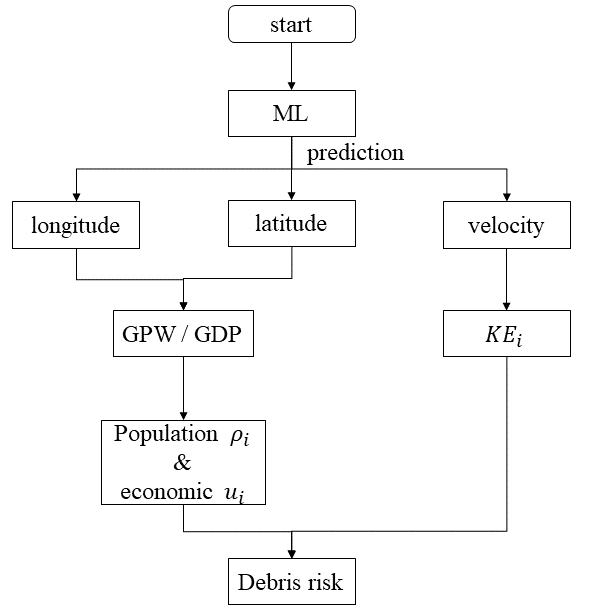}
	\caption{The flowchart of risk assessment}
	\label{fig:xfig021}
\end{figure}

Finally, according to the law of Bradford, the debris risk is divided into $5$ equal parts with a step size of $0.2$ ~\cite{law}. In the range of $0 ~ 1$, the value of spacecraft debris reentry risk is divided into $5$ intervals: $0 \sim 0.04 , 0.04 \sim 0.16 , 0.16 \sim 0.36 , 0.36 \sim 0.64 $, and  $0.64 \sim 1.0$ and named negligible risk, low risk, medium risk, high risk and very high risk, respectively. The flowchart of risk assessment as shown in Fig.~\ref{fig:xfig021}.

\section{Experiment}
\subsection{Dataset}
\begin{figure}[!htb] 
	\centering
	\includegraphics[width=0.7\textwidth]{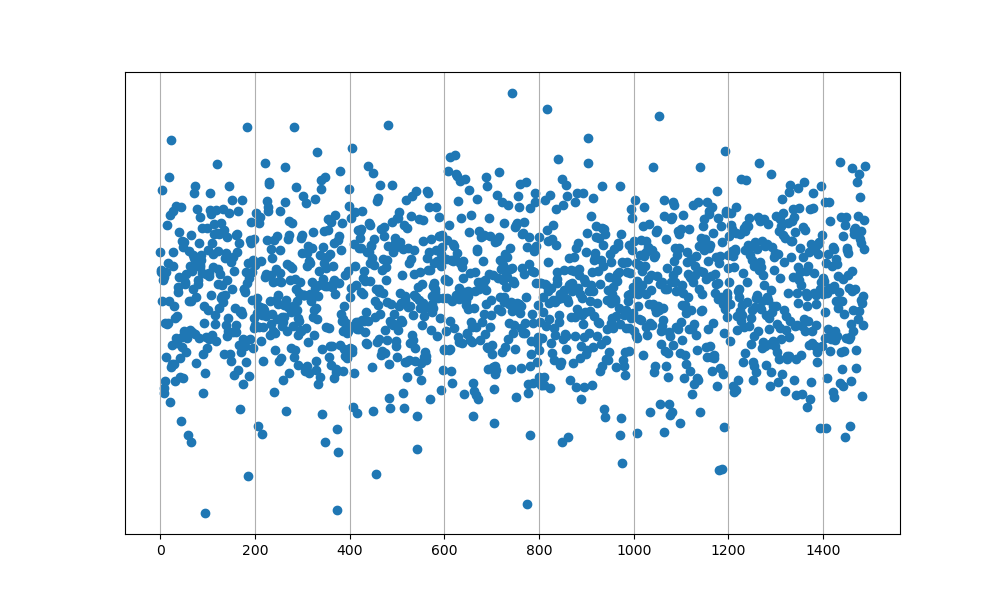}
	\caption{The data distribution of dataset in our paper}
	\label{fig:xfig999}
\end{figure}

The dataset in our paper consists of 1489 sample data points of spacecraft debris landings simulated by aerodynamic fusion trajectory. Of these data, 70\% is used as the training set and 30\% as the test set. In this paper, through statistical analysis and calculation of sample data, there are no obvious discrete points in the data set, as shown in Fig.~\ref{fig:xfig999}. Each sample consists of two parts:
\begin{enumerate}
	\item Data features. The ideal goal of machine learning is to generate the most efficient model with the fewest features. In complex data, there may be interactions between data features, and the value of one variable may seriously affect the importance of other variables. Therefore, the more features the model contains, the more complex the influence relationship between features is, the more sparse the data will be, thus making the model more sensitive to errors caused by variance. In this paper, Yellowbrick is used to analyze the importance of features on disintegrated data. Firstly, the recursive feature elimination method is used to select the number of features with the best score, as shown in Fig.~\ref{fig:xfig3_1}. Then, the importance of features is sorted and the parameters are selected as data features input into the machine learning model, as shown in Fig.~\ref{fig:xfig3_2}. Finally, six parameters were  obtained as data features input into the machine learning model, as shown in Table.~\ref{tab:table1}. 
	\begin{figure}[t]
		\centering
		\subfloat[]{\label{fig:xfig3_1}
			\includegraphics[width=0.45\textwidth]{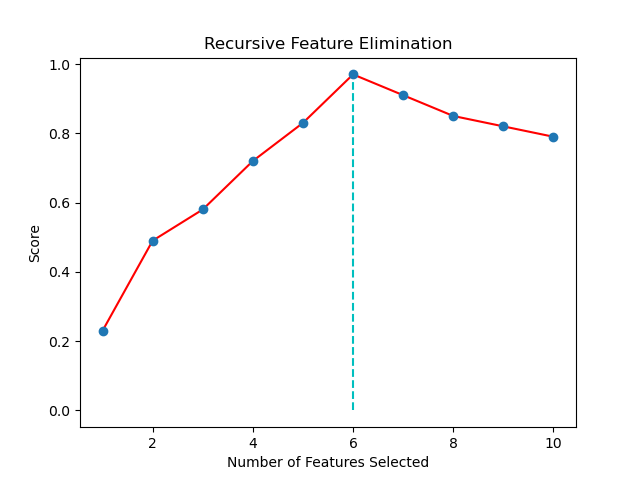}}
		\quad
		\subfloat[]{\label{fig:xfig3_2}
			\includegraphics[width=0.45\textwidth]{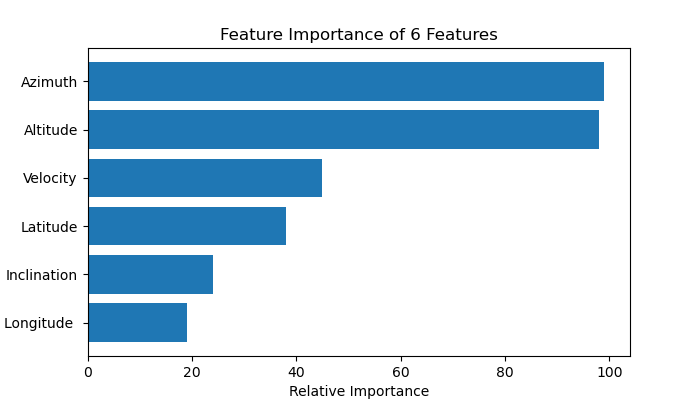}}
		\caption{The parameters are selected as data features. Figure (a) shows therecursive feature elimination method is used to select the number of features with the best score. Figure (b) shows the importance of features is sorted and parameters are selected as data features input into the machine learning model.}
		\label{fig:xfig3_0}
	\end{figure}
	
	\item Data labels. In this paper, firstly, the fragments are divided into several groups with simple geometric shapes by adopting a strategy similar to object-oriented method, and the mass distribution model of fragments is established. Then, the mass and number of each group after fragmentation generation are determined according to the fragment distribution model. Finally, the longitude, latitude and landing speed of the seven spacecraft during reentry were decomposed as data labels of the training model, as shown in Table.~\ref{tab:table2}.
\end{enumerate}
\begin{table}[h]
	\centering
	\caption{The table of data features}
	\label{tab:table1}
	\begin{tabular}{|c|c|c|c|c|c|}
		\hline
		\makecell[c]{Longitude of \\ ground coordinate \\ system origin (deg)} 
		&\makecell[c]{ Latitude of \\ ground  coordinate \\ system origin (deg)}
		&\makecell[c]{ Azimuth of ground \\ coordinate system (deg)}
		&\makecell[c]{Initial \\ altitude (m)} 
		&\makecell[c]{Initial \\ velocity (m/s)} 
		&\makecell[c]{Initial trajectory \\ inclination (deg)} 
		\\ \hline
	\end{tabular}
\end{table}

\begin{table}[h]
	\centering
	\caption{The table of data labels}
	\label{tab:table2}
	\begin{tabular}{|c|c|c|}
		\hline
		\makecell[c]{Longitude of \\ debris landing point (deg)} 
		&\makecell[c]{Latitude of \\ debris landing point (deg)}
		&\makecell[c]{Velocity of \\ debris landing point (m/s)}
		\\ \hline
	\end{tabular}
\end{table}

\subsection{Model Settings}
Our experiment uses the functions in scikit-learn and adjusts the parameters to obtain the highest accuracy~\cite{fp}. All regression algorithms use the MSE loss function, and the multilayer perceptron uses the Adam optimisation algorithm. The parameters of each algorithm are as follows:
\begin{enumerate}
	\item	There are 2 parameters in the SVR: $C$ and $epsilon$. $C$ is the regularization parameter, and the regularization intensity is inversely proportional to $C$. $Epsilon$ is the epsilon tube, which is defined as the tolerance without any loss to the error. When we set $C = 6.13$ and $epsilon = 5$, the model has the highest accuracy score.
	
	\item 	In the decision tree model, the maximum depth of the tree is $max\_ depth=5$.
	
	\item In the MLP model, the solution function of weight optimisation is $lbfgs$, the penalty parameter $\alpha = 1e-5$, and the maximum number of iterations $max\_iter = 500$.
\end{enumerate}

\subsection{Evaluation metric}
The performance evaluation metric of regression model mainly includes RMSE (square root error), MAE (mean absolute error), MSE (mean square error) and $R^2$ score. Noteworthy, when the dimensions are different, RMSE, MAE and MSE are difficult to measure the effect of the model. With that in mind, we use the $R ^ 2$ score as the accuracy evaluation metric in this paper. The $R ^ 2$ score reflects the proportion of the total variation of the dependent variable that can be explained by the independent variable through regression. A higher score represents a more accurate prediction result. When the score is negative, it means that the model prediction result is very bad or even worse than using the mean directly. The $R ^ 2$ score is calculated as follows:

\begin{equation}
	\label{equ:23}
	R^{2}=1- \frac{SS_{res}}{SS_{tot}}
\end{equation}

Where $SS_{tot}$ is the square sum of the difference between the target value and the mean of the data, and $SS_ {res}$ is the residual square sum of the difference between the target value and the prediction value. The calculation formula is shown in Eq.~\ref{equ:24}~\ref{equ:25}. Where $y_{i}$ and $\hat{y_{i}}$ are the target value and prediction value respectively,  $\overline{y}$ is the mean of all the target values.
\begin{equation}
	\label{equ:24}
	\begin{aligned}
		SS_{tot} = \sum_{i}(y_{i}- \overline{y})^2
	\end{aligned}
\end{equation}
\begin{equation}
	\label{equ:25}
	\begin{aligned}
		SS_{res} &= \sum_{i}(y_{i}-\hat{y}_{i})^2
		\\
		&=\sum_{i}e_{i}^2
	\end{aligned}
\end{equation}

\subsection{Results}
We calculate the $R ^ 2$ scores of the accuracy of three regression algorithms for the landing velocity, longitude and latitude of seven spacecraft fragments, as shown in Table.~\ref{tab:table3} \ref{tab:table4} \ref{tab:table5}. The results show that the decision tree regression algorithm is more accurate than most other regression algorithms in predicting the velocity, longitude and latitude of spacecraft debris landing points. The average longitude prediction accuracy of the seven fragments is $99.53\%$, and the highest is $99.68\%$. The average latitude prediction accuracy of the seven fragments is $99.32\%$, and the highest is $99.49\%$. The average velocity prediction accuracy of the seven fragments is $65.88\%$, and the highest is $72.53\%$. 
\begin{table}[h]
	\centering 
	\tabcolsep=0.6cm
	\renewcommand\arraystretch{2}
	\caption{The prediction accuracy of the longitude model. The best results are highlighted.}
	\label{tab:table3}
	\begin{tabular}{|c|c|c|c|}
	
		\hline
		
		\makecell[c]{Debris  number} 
		&\makecell[c]{SVR}
		&\makecell[c]{MLP}
		&\makecell[c]{DTR}
		\\ \hline
		\makecell[c]{1} 
		&\makecell[c]{81.54\%}
		&\makecell[c]{92.22\%}
		&\makecell[c]{\textbf{99.58\%}}
			\\ \hline
		\makecell[c]{2} 
		&\makecell[c]{81.92\%}
		&\makecell[c]{91.28\%}
		&\makecell[c]{\textbf{99.62\%}}
			\\ \hline
		\makecell[c]{3} 
		&\makecell[c]{81.49\%}
		&\makecell[c]{94.72\%}
		&\makecell[c]{\textbf{99.47\%}}
			\\ \hline
		\makecell[c]{4} 
		&\makecell[c]{82.03\%}
		&\makecell[c]{93.32\%}
		&\makecell[c]{\textbf{99.68\%}}
			\\ \hline
		\makecell[c]{5} 
		&\makecell[c]{81.42\%}
		&\makecell[c]{91.30\%}
		&\makecell[c]{\textbf{99.09\%}}
			\\ \hline
		\makecell[c]{6} 
		&\makecell[c]{81.78\%}
		&\makecell[c]{91.01\%}
		&\makecell[c]{\textbf{99.65\%}}
			\\ \hline
		\makecell[c]{7} 
		&\makecell[c]{81.99\%}
		&\makecell[c]{94.22\%}
		&\makecell[c]{\textbf{99.65\%}}
		\\ \hline
	\end{tabular}
\end{table}
\begin{table}[h]
	\centering
	\tabcolsep=0.6cm
	\renewcommand\arraystretch{2}
	\caption{The prediction accuracy of the latitude model. The best results are highlighted.}
	\label{tab:table4}
	\begin{tabular}{|c|c|c|c|}
		\hline
		\makecell[c]{Debris  number} 
		&\makecell[c]{SVR}
		&\makecell[c]{MLP}
		&\makecell[c]{DTR}
		\\ \hline
		\makecell[c]{1} 
		&\makecell[c]{96.23\%}
		&\makecell[c]{98.32\%}
		&\makecell[c]{\textbf{99.49\%}}
		\\ \hline
		\makecell[c]{2} 
		&\makecell[c]{96.97\%}
		&\makecell[c]{98.24\%}
		&\makecell[c]{\textbf{99.39\%}}
		\\ \hline
		\makecell[c]{3} 
		&\makecell[c]{96.05\%}
		&\makecell[c]{98.52\%}
		&\makecell[c]{\textbf{99.47\%}}
		\\ \hline
		\makecell[c]{4} 
		&\makecell[c]{96.99\%}
		&\makecell[c]{97.29\%}
		&\makecell[c]{\textbf{99.06\%}}
		\\ \hline
		\makecell[c]{5} 
		&\makecell[c]{96.01\%}
		&\makecell[c]{98.62\%}
		&\makecell[c]{\textbf{99.17\%}}
		\\ \hline
		\makecell[c]{6} 
		&\makecell[c]{96.97\%}
		&\makecell[c]{98.35\%}
		&\makecell[c]{\textbf{99.36\%}}
		\\ \hline
		\makecell[c]{7} 
		&\makecell[c]{96.99\%}
		&\makecell[c]{97.82\%}
		&\makecell[c]{\textbf{99.32\%}}
		\\ \hline
	\end{tabular}
\end{table}
\begin{table}[h]
	\centering
	\tabcolsep=0.6cm
	\renewcommand\arraystretch{2}
	\caption{The prediction accuracy of the velocity model. The best results are highlighted.}
	\label{tab:table5}
	\begin{tabular}{|c|c|c|c|}
		\hline
		\makecell[c]{Debris  number} 		&\makecell[c]{SVR}
		&\makecell[c]{MLP}
		&\makecell[c]{DTR}
		\\ \hline
		\makecell[c]{1} 
		&\makecell[c]{-0.38\%}
		&\makecell[c]{-10.99\%}
		&\makecell[c]{\textbf{63.81\%}}
		\\ \hline
		\makecell[c]{2} 
		&\makecell[c]{-24.09\%}
		&\makecell[c]{-33.95\%}
		&\makecell[c]{\textbf{65.90\%}}
		\\ \hline
		\makecell[c]{3} 
		&\makecell[c]{-43.05\%}
		&\makecell[c]{-9.64\%}
		&\makecell[c]{\textbf{72.53\%}}
		\\ \hline
		\makecell[c]{4} 
		&\makecell[c]{-11.94\%}
		&\makecell[c]{-7.43\%}
		&\makecell[c]{\textbf{60.94\%}}
		\\ \hline
		\makecell[c]{5} 
		&\makecell[c]{-64.68\%}
		&\makecell[c]{-9.11\%}
		&\makecell[c]{\textbf{66.42\%}}
		\\ \hline
		\makecell[c]{6} 
		&\makecell[c]{-50.83\%}
		&\makecell[c]{-9.78\%}
		&\makecell[c]{\textbf{65.23\%}}
		\\ \hline
		\makecell[c]{7} 
		&\makecell[c]{-35.99\%}
		&\makecell[c]{-19.64\%}
		&\makecell[c]{\textbf{66.32\%}}
		\\ \hline
	\end{tabular}
\end{table}

From the accuracy charts of different regression models, we find that the accuracy of the latitude model trained by the SVR and MLP is generally higher than that of the longitude model, but the accuracy of the latitude model trained by DTR is almost the same as that of the longitude model. According to the data distribution in our dataset, the reason is that the latitude range of the debris landing points is limited by orbital inclination. However, the longitude range is not limited.

\begin{figure}[!htb]
	\centering
	\subfloat[]{\label{fig:xfig004_1}
		\includegraphics[width=0.45\textwidth]{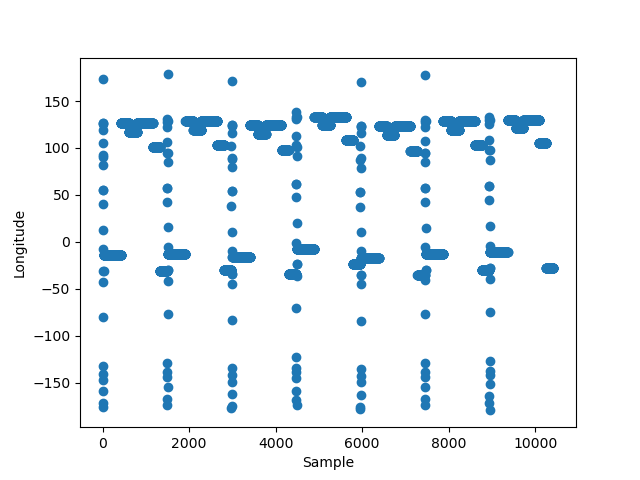}}
	\quad
	\subfloat[]{\label{fig:xfig004_4}
		\includegraphics[width=0.45\textwidth]{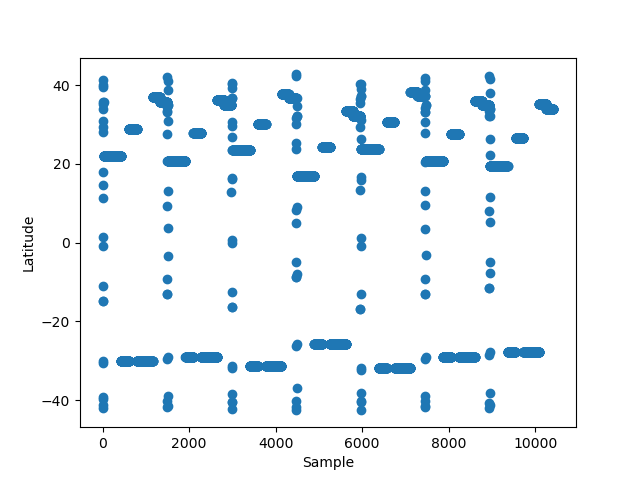}}
	\caption{Latitude and longitude distribution of samples in the dataset. Figures (a) show the distribution of longitude. Figure (b) the distribution of latitude.}
	\label{fig:xfig004}
\end{figure}

As shown in Fig.~\ref{fig:xfig004}, for the dataset in our paper, the orbital inclination is $42^{\circ}$, and the spacecraft can only reenter at latitude between $42^{\circ}S$ and $42^{\circ}N$. Therefore, the latitude ranges from $42^{\circ} S$ to $42^{\circ} N$, the longitude ranges from $180^{\circ} W$ to $180^{\circ} E$, and the value range of the latitude is smaller than that of the longitude. Therefore, the accuracy of the latitude is generally higher than longitude with the same amount of training. But the DTR divides the feature space into several units, and each unit has a specific output, which is not affected by the data distribution range.

\begin{figure}[b]
	\centering
	\subfloat[]{\label{fig:xfig4_1}
		\includegraphics[width=0.45\textwidth]{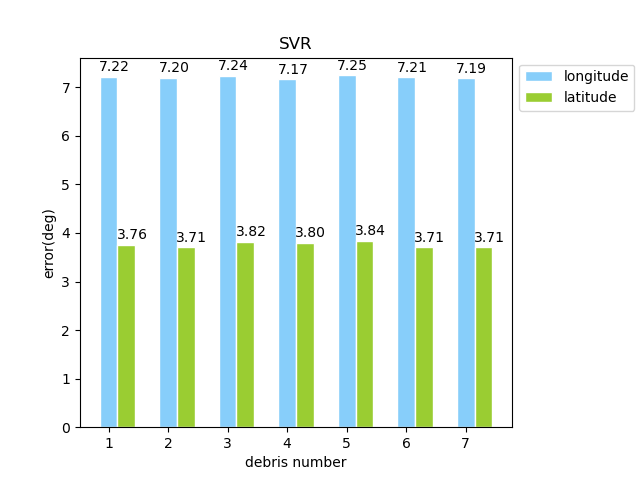}}
	\quad
	\subfloat[]{\label{fig:xfig4_2}
		\includegraphics[width=0.45\textwidth]{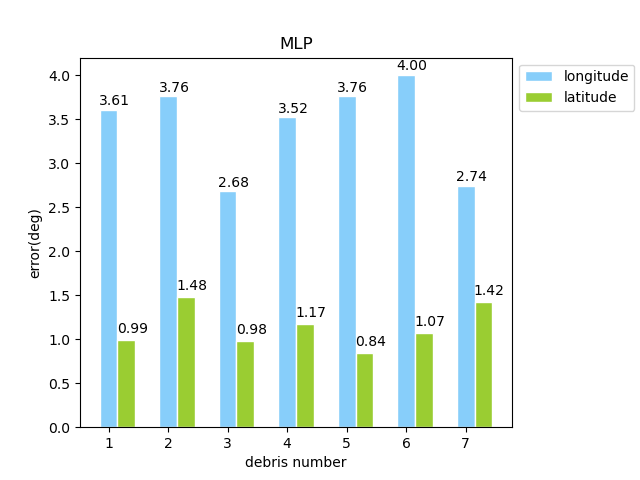}}
	\quad
	\subfloat[]{\label{fig:xfig4_3}
		\includegraphics[width=0.45\textwidth]{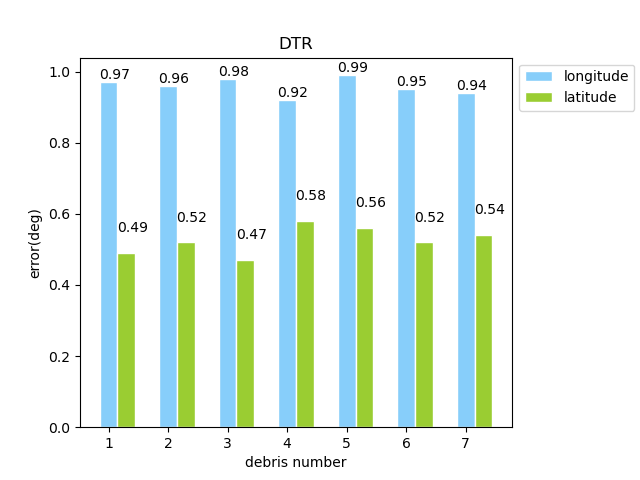}}
	\quad
	\subfloat[]{\label{fig:xfig4_4}
		\includegraphics[width=0.45\textwidth]{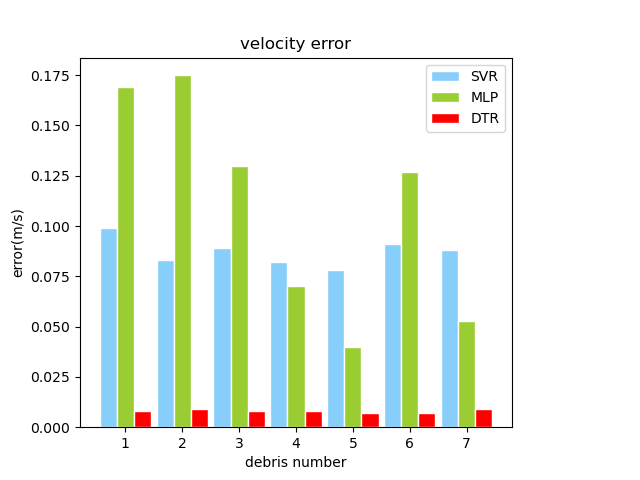}}
	\caption{The prediction error. Figures (a), (b), and (c) show the longitude and latitude errors predicted by the SVR, MLP and DTR machine learning regression algorithms, respectively. Figure (d) shows the comparison of the speed errors predicted by the SVR, MLP and DTR.}
	\label{fig:xfig4}
\end{figure}

In addition to evaluating the model with the $R^2$ score. In our paper, we calculate the difference between the prediction and the true value of the proposed three regression algorithms as the error, and the results are shown in Fig.~\ref{fig:xfig4}. The results show that the decision tree regression has the best performance among the three regression algorithms. The average longitude error of the seven spacecraft debris landing points is approximately $0.96^{\circ}$, the average latitude error is approximately $0.53^{\circ}$, and the average velocity error is $0.008m/s$. For all machine learning regression algorithms, because the range of longitude data in the dataset is larger than the range of latitude data, the longitude error of the spacecraft debris impact points is generally higher than the latitude error. Due to the small range of spacecraft debris landing velocities in the dataset, although the accuracy of the model is not high, the prediction error is still small. Taking the velocity prediction models of debris numbers 1 and 5 as examples, although the prediction accuracy of the landing velocity of debris number 1 is higher than 5 when using the SVR model, in our dataset, the landing speed scope of debris 5 is more concentrated, and the range is not large. Therefore, the model prediction results are also relatively concentrated in a small scope, and the prediction error is smaller for debris 5 than for debris 1.

\begin{figure}[!]
	\centering
	\subfloat[]{\label{fig:xfig5_1}
		\includegraphics[width=0.45\textwidth]{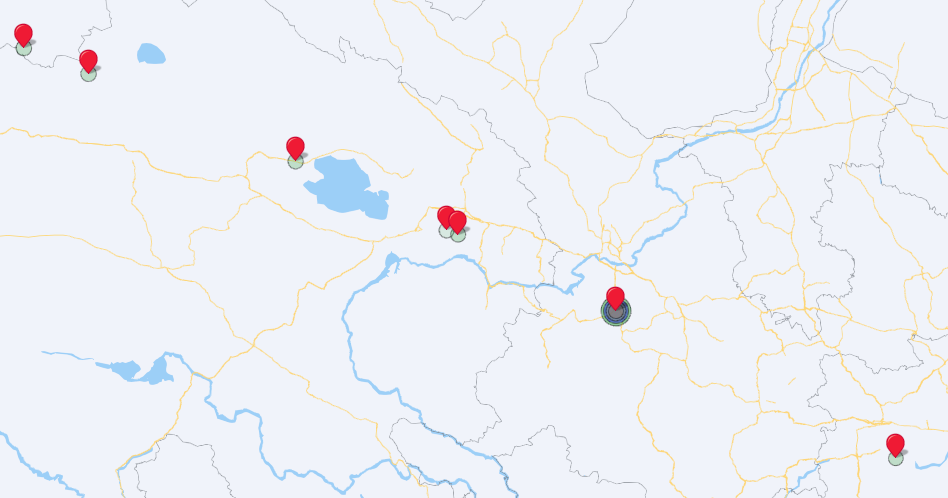}}
	\quad
	\subfloat[]{\label{fig:xfig5_2}
		\includegraphics[width=0.45\textwidth]{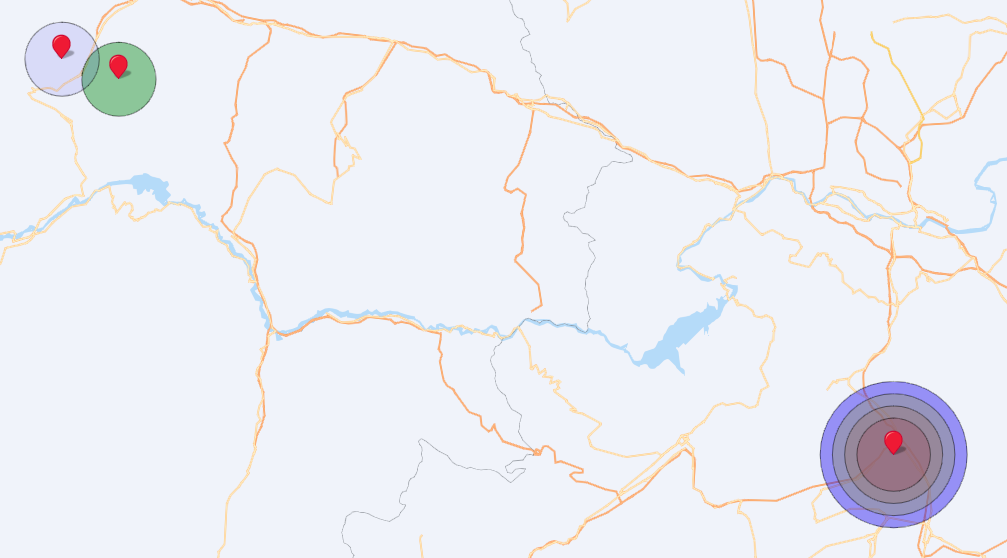}}
	\caption{The landing points of spacecraft debris. Figure (a) shows the landing map of the seven pieces of spacecraft debris. Figure (b) shows the degrees of danger of the spacecraft debris.}
	\label{fig:xfig5}
\end{figure}

On the basis of the reentry risk defined in our paper, the final degree of danger is calculated. According to the law of Bradford, the debris danger is divided into five degrees from low to high with a step size of $0.2$, which are named negligible risk, low risk, medium risk, high risk and very high risk, respectively. We use the five colours of green, blue, yellow, orange and red to respectively represent the five degrees of danger and then plot colour gradient concentric circles on the map with the spacecraft debris landing points as the centre points. The higher the degree of danger is, the larger the circle, which means a larger area of risk on the ground. The results are shown in Fig.~\ref{fig:xfig5}.

Finally, we calculate the running times of the three machine learning regression algorithms, including the training time required by the longitude and latitude prediction model of spacecraft debris landing points, the training time required by the velocity model and the total running time required by the program. In this paper, we calculate the means and standard deviations of the three regression algorithms after five trials each. Table~\ref{tab:table6} shows that the model training time and the total running time of the program using the decision tree regression algorithm have the lowest standard deviation, and the mean training time of the longitude and latitude prediction model is lower for the decision tree regression algorithm than the other two regression algorithms; furthermore, the mean velocity model training time and the mean total running time of the program using the SVR are the lowest. 

\begin{table}[]
	\centering
	\tabcolsep=0.6cm
	\renewcommand\arraystretch{2}
	\caption{The mean and standard deviation of the model training time and total program running time (seconds). The best results are highlighted.}
	\label{tab:table6}
	\begin{tabular}{|c|c|c|c|c|}
		\hline
		\makecell[c]{}
		&\makecell[c]{}
		&\makecell[c]{SVR}
		&\makecell[c]{MLP}
		&\makecell[c]{DTR}
		\\ \hline
		\multirow{2} * {\makecell{Longitude and latitude \\ prediction model training time}}
		&\makecell[c]{mean} 
		&\makecell[c]{0.8481}
		&\makecell[c]{15.0535}
		&\makecell[c]{\textbf{0.0146}}
		\\
		\cline{2-5}
		~&std & 0.0156 & 1.2231 & \textbf{0.003}
		\\ \hline
		
		\multirow{2} * {\makecell[c]{Velocity prediction model \\ training time}}
		&\makecell[c]{mean} 
		&\makecell[c]{\textbf{0.0197}}
		&\makecell[c]{9.5358}
		&\makecell[c]{0.0295}
		\\
		\cline{2-5}
		~&std & 0.0107 & 0.8293 & \textbf{0.0069} 
		\\ \hline
		
		\multirow{2} *{\makecell[c]{Total program run time}}
		&\makecell[c]{mean} 
		&\makecell[c]{\textbf{9.8294}}
		&\makecell[c]{35.5315}
		&\makecell[c]{10.5027}
		\\
		\cline{2-5}
		~&std & 1.2868 & 1.2103 & \textbf{0.5207}
		\\ \hline
	\end{tabular}
\end{table}

\section{Conclusion}
In this paper, we present a reetry risk and safety assessment of spacecraft debris based on machine learning. Compared with traditional physical modelling methods, the introduction of machine learning algorithms does not need to consider the re-entry process of intermediate spacecraft debris. And many experiments prove that the proposed method can obtain high accuracy prediction results in at least 15 seconds and make safety level warning more real-time. We compare the support vector regression model, the decision tree regression model and the multilayer perceptron regression model with three machine learning regression algorithms used to predict the spacecraft debris landing points according to the accuracy of the longitude, latitude and velocity. And we found that the decision regression tree had the highest prediction accuracy

What's more, we redefined the degrees of spacecraft reentry risk and the economy, population and kinetic energy of the debris landing points are taken as risk factors. Then, according to the law of Bradford, we divided the debris risk into 5 equal parts with a step size of 0.2, and formulated the degree of debris risk.
 In future work, we will expand the dataset size and then introduce deep learning methods to the prediction of the spacecraft reentry process.
\section*{Declarations}

\subsection*{Funding}
This research is supported by the National Key Research and Development Program of China under Grant No.2020YFC1523303?the National Natural Science Foundation of China under Grant No. 61672102, No.61073034, No. 61370064 and No. 60940032; the National Social Science Foundation of China under Grant No.BCA150050; the Program for New Century Excellent Talents in the University of Ministry of Education of China under Grant No. NCET-10-0239; the Open Project Sponsor of Beijing Key Laboratory of Intelligent Communication Software and Multimedia under Grant No. ITSM201493.

%
\subsection*{Conflicts of interest}
\noindent Not applicable

%

\bibliographystyle{unsrt}      
\bibliography{ref}   

%
%
\end{sloppypar}
\end{document}